\pdfoutput=1
\documentclass[11pt]{article}

\usepackage{acl}
\usepackage{geometry}
\usepackage{times}
\usepackage{latexsym}
\usepackage{amsmath}
\usepackage{amsthm}
\usepackage[psamsfonts]{amssymb}
\usepackage{graphicx}

\usepackage[T1]{fontenc}
\usepackage[utf8]{inputenc}
\usepackage{microtype}

\usepackage{url}
\usepackage{gb4e}
\usepackage[inference]{semantic}
\usepackage{stmaryrd}
\usepackage{natbib}
\bibpunct{(}{)}{;}{a}{,}{,}

\newcommand{\refex}[1]{(\ref{ex:#1})}

\newcommand{\refsec}[1]{Section~\ref{sec:#1}}

\newcommand{\refdef}[1]{Definition~\ref{def:#1}}

\newcommand{\refth}[1]{Theorem~\ref{th:#1}}

\newcommand{\reflem}[1]{Lemma~\ref{lem:#1}}

\newcommand{\reffig}[1]{Figure~\ref{fig:#1}}

\newcommand{\deft}[1]{\textit{#1}\index{#1}}

\newenvironment{centre}{\begin{center}$\begin{array}{c}}{\end{array}$\end{center}}

\newcommand{\rulesbox}[2][1]{
\hfil\begin{center}
\scalebox{#1}[#1]{\ensuremath{\begin{array}{l}
#2
\end{array}}}
\end{center}
}

\usepackage{fancybox}
\usepackage{dashbox}
\newenvironment{basicshadowbox}%
  {\medskip%
   \noindent%
   \begin{Sbox}%
   \begin{minipage}{.95\textwidth}%
   }{%
   \end{minipage}%
   \end{Sbox}%
   \shadowbox{\TheSbox}} 
\newenvironment{basicstackedbox}%
  {\medskip%
   \noindent%
   \begin{Sbox}%
   \begin{minipage}{.95\textwidth}%
   }{%
   \end{minipage}%
   \end{Sbox}%
   \lbox[1]{\fbox{\TheSbox}}
   }
  {\medskip%
   \noindent%
   \begin{Sbox}%
   \begin{minipage}{.95\textwidth}%
   }{
   \end{minipage}%
   \end{Sbox}%
   \dbox{\TheSbox}
   }
\cornersize{.2}
  {\medskip%
   \noindent%
   \begin{Sbox}%
   \begin{minipage}{.95\textwidth}%
   }{
   \end{minipage}%
   \end{Sbox}%
   \ovalbox{\TheSbox}
   }
  {\resetcounter{equation}%
   \begin{basicshadowbox}%
   \begin{definition}[#1]%
   }{%
   \end{definition}%
   \end{basicshadowbox}} 
  {\resetcounter{equation}%
   \begin{basicshadowbox}%
   \begin{axiom}[#1]%
   }{%
   \end{axiom}%
   \end{basicshadowbox}}
  {\resetcounter{equation}%
   \begin{basicstackedbox}%
   \begin{theorem}[#1]%
   }{
   \end{theorem}%
   \end{basicstackedbox}%
   }
  {\resetcounter{equation}%
   \begin{basicstackedbox}%
   \begin{lemma}[#1]%
   }{
   \end{lemma}%
   \end{basicstackedbox}%
   }
  {\resetcounter{equation}%
   \begin{basicshadowbox}%
   \noindent\textbf{\refdef{#2}}\ #1%
   }{%
   \end{basicshadowbox}%
   } 
  {\resetcounter{equation}%
   \begin{basicshadowbox}%
   \begin{axiom}[#1]%
   }{%
   \end{axiom}%
   \end{basicshadowbox}}
  {\resetcounter{equation}%
   \begin{basicstackedbox}%
   \noindent\textbf{\refth{#2}}\ #1%
   }{
   \end{basicstackedbox}%
   }
  {\resetcounter{equation}%
   \begin{basicstackedbox}%
   \noindent\textbf{\reflem{#2}}\ #1%
   }{
   \end{basicstackedbox}%
   }

\newcommand{\resetcounter}[1]{\setcounter{#1}{0}}

\usepackage{proof}
\usepackage{logic}
\usepackage{ccg}
\usepackage{dts}

\title{Is Japanese CCGBank empirically correct? \\
A case study of passive and causative constructions}

\author{Daisuke Bekki \\
  Ochanomizu University \\
  \texttt{bekki@is.ocha.ac.jp} \\\And
  Hitomi Yanaka \\
  The University of Tokyo \\
  \texttt{hyanaka@is.s.u-tokyo.ac.jp} \\}

\begin{document}
\maketitle
\begin{abstract}
The Japanese CCGBank serves as training and evaluation data for developing Japanese CCG parsers.  However, since it is automatically generated from the Kyoto Corpus, a dependency treebank, its linguistic validity still needs to be sufficiently verified.  In this paper, we focus on the analysis of passive/causative constructions in the Japanese CCGBank and show that, together with the compositional semantics of ccg2lambda, a semantic parsing system, it yields empirically wrong predictions for the nested construction of passives and causatives.
\end{abstract}

\begin{figure*}[t]
\rulesbox[.8]{
\ph[>]{
  \ccat{S}
  }{
  \clex{Taro-ga}{Taro-NOM}{\vT/(\vT\bs\np\f{ga})}
  &
  \ph[>]{
    \ccat{S\bs\np\f{ga}}
    }{
    \clex{Jiro-ni}{Jiro-DAT}{\vT/(\vT\bs\np\f{ni})}
    &
    \ph[<B_2]{
      \ccat{S\bs\np\f{ga}\bs\np\f{ni}}
      }{
      \ph[<]{
        \ccat{S\bs\np\f{ga}\bs\np\f{ni}}
        }{
        \clex{homera}{praise}{S\bs\np\f{ga}\bs\np\f{o}}
        &
        \clex{re}{passive}{S\bs\np\f{ga}\bs\np\f{ni}\bs (S \bs \np\f{ga}\bs\np\f{ni|o})}
      }
      &
      \clex{ta}{PST}{S \bs S}
      }
    }
  }
\\
\ph[>]{
  \ccat{S}
  }{
  \clex{Taro-ga}{Taro-NOM}{\vT/(\vT\bs\np\f{ga})}
  &
  \ph[>]{
    \ccat{S\bs\np\f{ga}}
    }{
    \clex{Jiro-o}{Jiro-ACC}{\vT/(\vT\bs\np\f{o})}
    &
    \ph[<B_2]{
      \ccat{S\bs\np\f{ga}\bs\np\f{ni|o}}
      }{
      \ph[<]{
        \ccat{S\bs\np\f{ga}\bs\np\f{ni|o}}
        }{
        \clex{hasira}{run}{S\bs\np\f{ga}}
        &
        \clex{se}{cause}{S\bs\np\f{ga}\bs\np\f{ni|o}\bs (S \bs \np\f{ga})}
        }
      &
      \clex{ta}{PST}{S \bs S}
      }
    }
  }
}
\caption{Syntactic structures of \refex{P1} and \refex{C1} in \citet{Bekki2010CCGbook}}\label{fig:P1C1trees}
\end{figure*}

\section{Introduction}
\label{sec:Introduction}%

The process of generating wide-coverage syntactic parsers from treebanks was established in the era of probabilistic context-free grammar (CFG) parsers in the 1990s.  However, it was believed at that time that such an approach did not apply to linguistically-oriented formal syntactic theories.  The reason was that formal syntactic theories were believed to be too inflexible to exhaustively describe the structure of real texts.  This misconception was dispelled by the theoretical development of formal grammars and the emergence of linguistically-oriented treebanks.%
\footnote{%
To mention a few, the LinGO Redwoods treebank~\citep{LinGO2002} contains English sentences annotated with Head-driven Phrase Structure Grammar (HPSG) parse trees.  The INESS treebank repository~\citep{INESS2012} offer Lexical Functional Grammar (LFG) treebanks such as The ParGram Parallel Treebank (ParGramBank)~\citep{ParGramBank2013}, which provides ten typologically different languages.
} %
In particlar, Combinatory Categorial Grammar (CCG) \citep{Steedman1996,Steedman2000} and  CCGbank~\citep{CCGbank2005} gave rise to the subsequent developments of CCG parsers such as C\&C parser~\citep{ClarkCurran2007} and EasyCCG parser~\citep{LewisSteedman2014}, and proved that wide-coverage CCG parsers could be generated from treebanks in a similar process to probablistic CFG parsers.

This trend has also impacted research on Japanese syntax and parsers. \citet{Bekki2010CCGbook} revealed that CCG, as a syntactic theory, enables us to provide a wide-coverage syntactic description of the Japanese language.  It motivated the development of the Japanese CCGBank \citep{Uematsu+2013}, followed by Japanese CCG parsers such as Jigg~\citep{NojiMiyao2016} and depccg~\citep{Yoshikawa2017}.

The difficulty in developing the Japanese CCGBank lay in the absence of CFG treebanks for the Japanese language at that time.%
\footnote{%
Recently, a large-scale CFG treebank for the Japanese language is available as a part of NINJAL parsed corpus of modern Japanese \url{https://npcmj.ninjal.ac.jp/}, and there is also an attempt to generate a treebank of better quality by using it \citep{Kubota+2020}.  However, the questions of what is empirically problematic about the Japanese CCGBank and, more importantly, why it is, remain undiscussed.  
 The importance of answering these questions as we do in this paper is increasing, given that attempts to generate a CCGBank from a dependency corpus such as Universal Dependency are still ongoing (cf. \citet{Tran2022}).}
While CCGbank was generated from the Penn Treebank, which is a CFG treebank, the only large-scale treebank available for Japanese was the Kyoto Corpus\footnote{\url{https://github.com/ku-nlp/KyotoCorpus}}, which is a dependency tree corpus, from which \citet{Uematsu+2013} attempted to construct a Japanese CCGBank by automatic conversion.

The syntactic structures of CCG have more elaborated information than those of CFG, such as argument structures and syntactic features.  Thus, it is inevitable that a dependency tree, which has even less information than that of CFG, must be supplemented with a great deal of linguistic information.  \citet{Uematsu+2013} had to guess them systematically, which is not an obvious process, and ad-hoc rules had to be stipulated in many places to accomplish it, including the ``passive/causative suffixes as $S \bs S$ analysis,'' which we will discuss in \refsec{ccgbank}.

Since CCGBank serves as both training and evaluation data for CCG parsers, syntactic descriptions in CCGBank set an upper bound on CCG parser performance, which inherit any empirical fallacies in CCGBank: thus the validity of the syntactic structures in CCGBank is important.  However, little research from the perspective of formal syntax has been conducted regarding the adequacy of syntactic structures contained in treebanks.

This paper aims to assess the syntactic structures exhibited by the Japanese CCGbank from the viewpoint of theoretical linguistics.  Specifically, we focus on the syntax and semantics of case alternation in passive and causative constructions in Japanese, a linguistic phenomenon analyzed differently in the standard Japanese CCG and CCGBank, and show that the syntactic analysis of the Japanese CCGBank contains empirical fallacies.

\begin{figure*}[t]
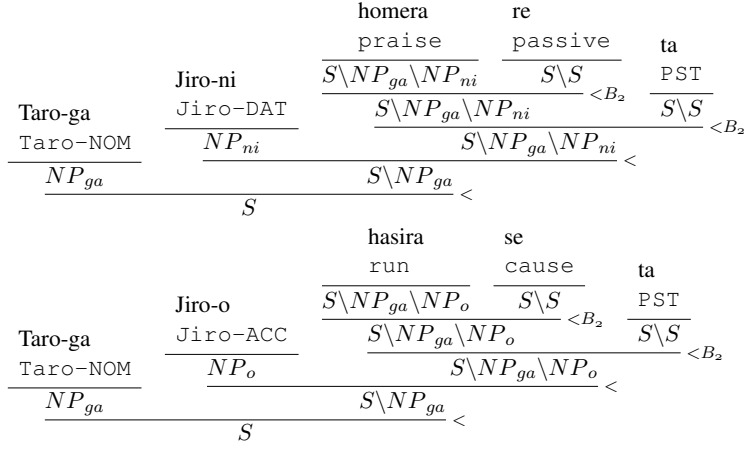

\rulesbox[.8]{
\ph[<]{
  \ccat{S}
  }{
  \clex{Taro-ga}{Taro-NOM}{\np\f{ga}}
  &
  \ph[<]{
    \ccat{S\bs\np\f{ga}}
    }{
    \clex{Jiro-ni}{Jiro-DAT}{\np\f{ni}}
    &
    \ph[<B_2]{
      \ccat{S\bs\np\f{ga}\bs\np\f{ni}}
      }{
      \ph[<B_2]{
        \ccat{S\bs\np\f{ga}\bs\np\f{ni}}
        }{
        \clex{homera}{praise}{S\bs\np\f{ga}\bs\np\f{ni}}
       &
        \clex{re}{passive}{S \bs S}
        }
      &
      \clex{ta}{PST}{S \bs S}
      }
    }
  } 
\\
\ph[<]{
  \ccat{S}
  }{
  \clex{Taro-ga}{Taro-NOM}{\np\f{ga}}
  &
  \ph[<]{
    \ccat{S\bs\np\f{ga}}
    }{
    \clex{Jiro-o}{Jiro-ACC}{\np\f{o}}
    &
    \ph[<B_2]{
      \ccat{S\bs\np\f{ga}\bs\np\f{o}}
      }{
      \ph[<B_2]{
        \ccat{S\bs\np\f{ga}\bs\np\f{o}}
        }{
        \clex{hasira}{run}{S\bs\np\f{ga}\bs\np\f{o}}
        &
        \clex{se}{cause}{S \bs S}
        }
      &
      \clex{ta}{PST}{S \bs S}
      }
    }
  }
}
\caption{Syntactic structures of \refex{P1} and \refex{C1} in CCGBank}\label{fig:P1C1ccg2lambda}
\end{figure*}

\section{Passive and Causative Constructions in Japanese}
\label{sec:PassiveCausative}%
We first present some empirical facts about Japanese passives and causatives and how they are described in the standard Japanese CCG~\citep{Bekki2010CCGbook}.
\textit{Ga}-marked noun phrases (henceforth \np\f{ga}) in passive sentences correspond to \textit{ni}-marked noun phrases (henceforth \np\f{ni}) or \textit{o}-marked noun phrases (henceforth \np\f{o}) in the corresponding active sentences, which is expressed as \refex{P1} in the form of inferences.
\begin{exe}
  \ex\label{ex:P1} 
  \gll Taro-ga Jiro-ni homera-re-ta $\Longrightarrow$ Jiro-ga Taro-o home-ta \\
       Taro-NOM Jiro-DAT praise-passive-PST {} Jiro-NOM Taro-ACC praise-PST \\
  \trans (trans.) `Taro is praised by Jiro.' $\Longrightarrow$ `Jiro praised Taro.'
\end{exe}

Next, \np\f{ni} or \np\f{o} in causative sentences correspond to \np\f{ga} in the corresponding active sentences, which is also expressed in the form of inference as in \refex{C1}.
\begin{exe}
  \ex\label{ex:C1} 
  \gll Taro-ga Jiro-\{ni$|$o\} hasira-se-ta  $\Longrightarrow$ Jiro-ga hasit-ta \\
       Taro-NOM Jiro-\{DAT$|$ACC\} run-causative-PST {} Jiro-NOM run-PST \\
  \trans (trans.) `Taro made Jiro run.' $\Longrightarrow$ `Jiro run.'
\end{exe}

\noindent According to \citet{Bekki2010CCGbook}, the syntactic structure of the left-side sentences of \refex{P1} and \refex{C1} are as shown in \reffig{P1C1trees}.

For simplicity (omitting analysis of tense, etc.), let us assume that the semantic representations of \textit{Taro-ga}, \textit{Jiro-\{ni{$|$}o\}}, \textit{homera}, \textit{hasira}, and \textit{ta} are respectively defined as 
$\LAM[P]{P(\pred{t})}$,$\LAM[P]{P(\pred{j})}$,
$\LAM[y]\LAM[x]\LAM[k]\dSigmaf[e]{ev}{\pred{praise}(e,x,y) \times ke}$,
$\LAM[x]\LAM[k]\dSigmaf[e]{ev}{\pred{run}(e,x) \times ke}$,
$\mathit{id}$
by using event semantics \citep{Davidson1967} with continuations \citep{Chierchia1995} in terms of DTS (dependent type semantics)~\citep{BekkiMineshima2017}, where $\mathit{id}$ is the identity function and $\mathit{ev}$ is the type for events.

Then the core of the analysis of case alternation is to define semantic representations of the passive suffix \textit{re} and the causative suffix \textit{se}, homomorphically to their syntactic categories, as
\begin{exe}
\ex\label{ex:re} Passive suffix \textit{re}: $\LAM[P]\LAM[y]\LAM[x]{Pxy}$
\ex\label{ex:se} Causative suffix \textit{se}:
\sn $\LAM[P]\LAM[y]\LAM[x]\LAM[k]Py(\LAM[e]\pred{cause}(e,x) \times ke)$
\end{exe}

\noindent In words, both suffixes \textit{know} the argument structure of its first argument, namely, the verb.  In passive constructions, \np\f{ga} corresponds to the \np\f{o} or \np\f{ni}, and \np\f{ni} corresponds to \np\f{ga}, in their active counterparts.  In causative constructions, \np\f{ni|o} corresponds to \np\f{ga} in their active counterparts.  Assuming the event continuation $k$ is replaced by the term $\LAM[e]\top$ at the end of the semantic composition (where $\top$ is an enumeration type with only one proof term and plays the role of ``true''), the semantic composition ends up in the following representations for the left-side sentences of \refex{P1} and \refex{C1}, respectively.
\begin{exe}
\ex $\dSigmaf[e]{ev}{\pred{praise}(e,\pred{j},\pred{t}) \times \top}$
\ex\label{ex:sem8} $\dSigmaf[e]{ev}{\pred{run}(e,\pred{j}) \times \pred{cause}(e,\pred{t}) \times \top}$
\end{exe}

These respectively entail the right-side sentences of \refex{P1} and \refex{C1}, the semantic representations of which are \refex{8} and \refex{9} respectively, so the inferences \refex{P1} and \refex{C1} are correctly predicted.
\begin{exe}
\ex\label{ex:8} $\dSigmaf[e]{ev}{\pred{praise}(e,\pred{j},\pred{t}) \times \top}$
\ex\label{ex:9} $\dSigmaf[e]{ev}{\pred{run}(e,\pred{j}) \times \top}$
\end{exe}

The validity of this analysis can be verified by inference data on various constructions including passives and causatives. In particular, causatives can be nested in passives in Japanese, as in \refex{N1}.
\begin{exe}
  \ex\label{ex:N1} 
  \gll Jiro-ga Taro-ni hasira-sera-re-ta $\Longrightarrow$ Taro-ga Jiro-o hasira-se-ta \\
       Jiro-NOM Taro-DAT run-causative-passive-PST {} Taro-NOM Jiro-ACC run-causative-PST \\
  \trans (lit.) `Jiro was made run by Taro.' $\Longrightarrow$ `Taro made Jiro run.'
\end{exe}

The constituent \textit{hasira-sera-re} is the passivization of \textit{hasira-sera}, the matrix predicate of the left-side sentence of \refex{C1} that is equivalent to the right-side sentence of \refex{N1}, and thus also entails the right-side sentence of \refex{C1}. The semantic representation of \textit{hasira-sera-re} is obtained by a functional application of \refex{se} to the semantic representation of \textit{hasira}, followed by a functional application of \refex{re}, as follows.
\begin{exe}
\ex\label{ex:sera}
$\LAM[y]\LAM[x]\LAM[k]{}$
\sn $\dSigmaf[e]{ev}{\pred{run}(e,x)\times\pred{cause}(e,y)\times ke}$
\end{exe}

Therefore, the semantic representation of the left-side of \refex{N1} is 
$\dSigmaf[e]{ev}{\pred{run}(e,\pred{j})\times\pred{cause}(e,\pred{t})\times\top}$,
which is equal to \refex{sem8}, so it is correctly predicted to entail the left and right-sides of \refex{C1}. Thus, the passive/causal analysis in \citet{Bekki2010CCGbook} robustly predicts and explains the process from syntactic structures to semantic representations, and inferences.

\section{ccg2lambda and the $S \bs S$ analysis}
\label{sec:ccgbank}%

Analysis using a compositional semantic system ccg2lambda~\cite{MartinezGomez2016} relies on the syntactic structures output by the Japanese CCG parsers Jigg or depccg. 
As mentioned in \refsec{Introduction}, the output of these CCG parsers depends on Japanese CCGBank. In Japanese CCGBank, the lexical assignments for the left-side sentences of \refex{P1} and \refex{C1} are as shown in \reffig{P1C1ccg2lambda}, in which both the passive suffix \textit{re} and the causative suffix \textit{se} have the syntactic category $S \bs S$.

Let us glance over how non-passive sentences in CCGBank are semantically analyzed. The semantic representation of the two-place predicate \textit{homera} is given as follows (slightly simplified).
\begin{exe}
\ex\label{ex:homeru2} $\lambda Q_2 Q_1 C_1 C_2 K.$
\sn $Q_1(\LAM[x_1] Q_2(\LAM[x_2] \exists e (K(\pred{praise},e)$ 
\sn $\&\ C_1(x_1,e,\pred{Ag})\ \&\ C_2(x_2,e,\pred{Th}))))$
\end{exe}

This appears to be considerably more complex than that of \textit{homera} in the previous section. 
This is because in ccg2lambda, the relations between $\pred{Ag}$ (=Agent) and $x_1,e$ and between $\pred{Th}$ (=Theme) and $x_2,e$ are relativized by the higher-order variables $C_1,C_2$.  After taking an \np\f{ni} for $Q_2$ and an \np\f{ga} for $Q_1$ to become the constituent of syntactic category $S$, ccg2lambda applies to it the function $\lambda S. S(\lambda x e T.(T(e) = x), \lambda x e T.(T(e) = x), \mathit{id})$.  This causes $\lambda x e T.(T(e) = x)$ to be assigned to $C_1$ and $C_2$ to specify $\pred{Ag}(e) = x_1$ and $\pred{Th}(e) = x_2$, and $\mathit{id}$ to be assigned to $K$. Assuming $\LAM[P]P(\pred{t})$ and $\LAM[P]P(\pred{j})$ for the semantic representations of \textit{Taro-ga} and \textit{Jiro-ni}, the semantic representation of the right-side of \refex{P1} is obtained as \refex{ccg2lambdaA}, which is a standard neo-Davidsonian semantic representation \citep{Parsons1990} for \textit{Jiro praised Taro}.
\begin{exe}
\ex\label{ex:ccg2lambdaA} $\exists e (\pred{praise}(e)\ \&\ \pred{Ag}(e)=\pred{j}\ \&\ \pred{Th}(e)=\pred{t})$
\end{exe}

By contrast, in ccg2lambda the semantic representation of \textit{re} is overwritten by the \textit{semantic template} as 
\begin{exe}
\ex\label{ex:ccg2lambdaRe} \scalebox{.9}{$\lambda Q_2 Q_1 C_1 C_2 K. V(Q_2,Q_1,$}
\sn \scalebox{.9}{$\lambda x_1 e T. C_1(x_1,e,\pred{Th}),\lambda x_2 e T. C_2(x_2,e,\pred{Ag}),K)$}
\end{exe}

$V$ is instantiated by the semantic representation of the adjacent transitive verb (=\textit{homera} in this case). The semantic representation of \textit{homera-re} thus becomes
\begin{exe}
\ex \scalebox{.9}{$\lambda Q_2 Q_1 C_1 C_2 K. Q_1(\LAM[x_1] Q_2(\LAM[x_2] \exists e (K($}
\sn \scalebox{.9}{$\pred{praise},e)\ \&\ C_1(x_1,e,\pred{Th})\ \&\ C_2(x_2,e,\pred{Ag}))))$}
\end{exe}

That is, the semantic roles received by $C_1,C_2$ in \refex{homeru2} are discarded, and instead $C_1$ is given \pred{Th} and $C_2$ is given \pred{Ag}. By applying $\LAM[P]P(\pred{t})$, $\LAM[P]P(\pred{j})$, and $\lambda S. S(\lambda x e T.(T(e)=x), \lambda x e T.(T(e)=x), \mathit{id})$ sequentially, the left-side of \refex{P1} becomes
\begin{exe}
\ex\label{ex:ccg2lambdaB} $\exists e (\pred{praise}(e)\ \&\ \pred{Ag}(e)=\pred{j}\ \&\ \pred{Th}(e)=\pred{t})$
\end{exe}
Because this is the same as \refex{ccg2lambdaA}, the inference \refex{P1} is correctly predicted. Similarly, for causative suffixes, given a semantic template
\begin{exe}
\ex\label{ex:ccg2lambdaSe} \scalebox{.9}{$\lambda Q_2 Q_1 C_1 C_2 K. V(Q_2,Q_1,\lambda x_1 e T.$}
\sn \scalebox{.9}{$C_1(x_1,e,\pred{Cause}),\lambda x_2 e T. C_2(x_2,e,\pred{Ag}),K)$}
\end{exe}
the semantic representation of \textit{hasira-se} is obtained as follows.
\begin{exe}
\ex\label{ex:ccg2lambdaHasirase} \scalebox{.9}{$\lambda Q_2 Q_1 C_1 C_2 K.Q_1(\LAM[x_1] Q_2(\LAM[x_2]K(\pred{run},e)$}
\sn \scalebox{.9}{$\exists e (\ \&\ C_1(x_1,e,\pred{Cause})\ \&\ C_2(x_2,e,\pred{Ag}))))$}
\end{exe}

Thus, the left-side of the \refex{C1} will be
\begin{exe}
\ex\label{ex:ccg2lambdaC} $\exists e (\pred{run}(e)\ \&\ \pred{Cause}(e)=\pred{t}\ \&\ \pred{Ag}(e)=\pred{j})$
\end{exe}

\noindent which entails $\exists e (\pred{run}(e)\&\ \pred{Ag}(e)=\pred{j})$, the right-side of \refex{C1}, so the inference \refex{C1} is also correctly predicted.

However, this analysis produces incorrect predictions for nesting: the semantic representation of \textit{hasira-sera-re} is obtained by applying \refex{ccg2lambdaRe} to \refex{ccg2lambdaHasirase}, which ends up in \refex{ccg2lambdaHasiraserare}.
\begin{exe}
\ex\label{ex:ccg2lambdaHasiraserare} $\lambda Q_2 Q_1 C_1 C_2 K. Q_1(\LAM[x_1] Q_2(\LAM[x_2] \exists e( $
\sn \scalebox{.9}{$K(\pred{run},e)\ \&\ C_1(x_1,e,\pred{Th})\ \&\ C_2(x_2,e,\pred{Ag}))))$}
\end{exe}

Notice that \refex{ccg2lambdaHasiraserare} is identical to the one obtained by applying passive suffix \textit{re} directly to \textit{hasira} (i.e., \textit{hasira-re}). From here, neither the right-side of \refex{N1} = \refex{C1} nor the left-side of \refex{C1} is implied.

This error occurs because the passive suffix globally assumes that the first argument is given \pred{Th} and the second argument is given \pred{Ag}. On the contrary, the nesting example shows that the passive suffix connects the first and the second arguments in matrix with the second and the first argument of the verb, respectively. Capturing this behaviour of the passive suffix requires the verb's first and second arguments to be accessible from the syntax and the semantics of the passive suffix, which means the syntactic category of the passive suffix should be exactly $S\bs\np\f{ga}\bs\np\f{ni}\bs(S\bs\np\f{ga}\bs\np\f{ni|o})$.

\section{Conclusion}
In this paper, we showed that the syntactic analysis of Japanese CCGBank together with the semantic analysis of ccg2lambda produces false predictions for passive and causative nesting, which means that the current syntactic analysis of passive/causative constructions in Japanese CCGBank does not have a semantic support that correctly predicts the inferences such as \refex{P1}, \refex{C1}, and \refex{N1}. In other words, the claim that \textit{re} and \textit{se} have the syntactic category $S \bs S$ cannot be maintained.
Since the standard analysis described in \refsec{PassiveCausative} correctly explains all of those inferences, the burden of proof is clearly on the CCGBank side.

An important implication of this paper is that there is a need for outreach to the linguistic community, where not all linguists regard a treebank as an output of a linguistic analysis. We suggest that we should treat treebanks as outputs of some linguistic analyses and try to provide counterexamples in this way in order to keep treebanks and also the subsequent development of syntactic parsers sound from the linguistic perspective.

\section*{Acknowledgements}
We thank the anonymous reviewers for their comments and suggestions.
This work was partially supported by JST CREST Grant Number JPMJCR20D2, Japan, and JSPS KAKENHI Grant Number JP20K19868, Japan.

\bibliography{MyLibraryX8.bib}
\bibliographystyle{acl_natbib}

\end{document}